\documentclass[conference]{IEEEtran}
\IEEEoverridecommandlockouts
\usepackage{cite}
\usepackage{afterpage}
\usepackage{amsmath,amssymb,amsfonts}
\usepackage{algorithmic}
\usepackage{algorithm}
\usepackage[french,vlined,ruled,algo2e]{algorithm2e}
\usepackage[ruled]{algorithm2e}
\usepackage{graphicx}
\usepackage{textcomp}
\usepackage{xcolor}
\usepackage{multirow}
\def\BibTeX{{\rm B\kern-.05em{\sc i\kern-.025em b}\kern-.08em
    T\kern-.1667em\lower.7ex\hbox{E}\kern-.125emX}}
\DeclareMathOperator*{\argmin}{arg\,min}
\newcommand\blfootnote[1]{%
  \begingroup
  \renewcommand\thefootnote{}\footnote{#1}%
  \addtocounter{footnote}{-1}%
  \endgroup
}

\begin{document}

\title{Joint Multi-View Collaborative Clustering}

\author{\IEEEauthorblockN{Yasser KHALAFAOUI*}
\IEEEauthorblockA{\textit{R\&D department} \\
\textit{ALTECA}\\
Massy, France \\
mykhalafaoui@alteca.fr}
\and
\IEEEauthorblockN{Basarab MATEI}
\IEEEauthorblockA{\textit{LIPN - CNRS UMR 7030} \\
\textit{Sorbonne Paris Nord University}\\
Villetaneuse, France \\
matei@lipn.univ-paris13.fr}
\and
\IEEEauthorblockN{Nistor GROZAVU}
\IEEEauthorblockA{\textit{ETIS - CNRS UMR 8051} \\
\textit{CY Cergy Paris University}\\
Cergy, France \\
nistor.grozavu@cyu.fr}
\and
\IEEEauthorblockN{Martino LOVISETTO}
\IEEEauthorblockA{\textit{R\&D department} \\
\textit{ALTECA}\\
Lyon, France \\
mlovisetto@alteca.fr}

}

\maketitle
\begin{abstract}
Data is increasingly being collected from multiple sources and described by multiple views. These multi-view data provide richer information than traditional single-view data. Fusing the former for specific tasks is an essential component of multi-view clustering. Since the goal of multi-view clustering algorithms is to discover the common latent structure shared by multiple views, the majority of proposed solutions overlook the advantages of incorporating knowledge derived from horizontal collaboration between multi-view data and the final consensus. To fill this gap, we propose the Joint Multi-View Collaborative Clustering (JMVCC) solution, which involves the generation of basic partitions using Non-negative Matrix Factorization (NMF) and the horizontal collaboration principle, followed by the fusion of these local partitions using ensemble clustering. Furthermore, we propose a weighting method to reduce the risk of negative collaboration (i.e., views with low quality) during the generation and fusion of local partitions. The experimental results, which were obtained using a variety of data sets, demonstrate that JMVCC outperforms other multi-view clustering algorithms and is robust to noisy views.
\end{abstract}

\blfootnote{* Corresponding author}
\section{Introduction}
Data collections became very diverse in the recent decade \cite{b1} as a result of the advent of multi-modal data sets, multi-view data sets (i.e. the same data sample described in different ways), and distributed data. For example, captions and textual tags are frequently used to describe images, and some news are covered by several media organisations and in multiple languages. Given that these different representations frequently provide compatible and complimentary information \cite{b2}, it is essential to extract intrinsic information from these multi-view data sets rather than focusing on standard single-views. One of the most important aspects of learning from multi-view data sets is to use the extracted knowledge from each view to deal with outliers and noisy characteristics \cite{b3}.\par
The task of finding intrinsic information across multiple views, known as multi-view clustering, has received a lot of attention \cite{b4,b5,b6}. The objective of multi-view clustering is to leverage heterogeneous information from different views to create a high quality clustering result. There are four types of multi-view clustering approaches in the literature. \textit{Multi-view graph clustering} is a set of methods that finds a fusion graph across all views and then applies clustering techniques to the fusion graph to produce the clustering result. Wang et al. \cite{b7} created a nearest neighbor graph to find an encoding for the associated manifold information. Then, to determine the best intrinsic manifold, they built a multiple graph ensemble regularization framework. Cross diffused matrix alignment based on feature selection is a technique that Wei et al. \cite{b8} proposed for choosing features for each view while doing alignment on a cross diffused matrix. The final clustering results were then obtained via co-regularized spectral clustering on these chosen features \cite{b9}. \textit{Multi-kernel learning} uses predetermined kernels related to several views that are subsequently combined either linearly or non-linearly to enhance clustering performance \cite{b10}. Based on the minimizing-disagreement methodology, Su et al. \cite{b11} developed a novel kernel combination method. They created a multi-partite graph to infer a kernel, which was then applied to spectral clustering. \textit{Multi-task multi-view clustering} enhance clustering performance by assigning one or more tasks to each view, distributing inter-task knowledge, and making use of connections between many tasks and views. Wang et al. \cite{b12} investigated multi-view spectral clustering using a multi-objective framework (viewed as a multi-task), which is addressed by Pareto optimization. \textit{Collaborative clustering} is a co-training strategy used to handle multi-view data. It bootstraps the clustering of multiple views by leveraging the knowledge extracted from each view \cite{b13}. When this procedure is applied iteratively, the clustering results of all views tend to converge. Jiang et at. \cite{b14} established a co-regularized Probabilistic Latent Semantic Analysis (PLSA) model for multi-view co-clustering.\par
However, some views with irrelevant or noisy features may negatively impact the common space and decrease clustering quality. Furthermore, as the number of views grows, there is less common area shared by all views.\par
In recent years, Non-negative Matrix Factorization has gained a lot of attention and has been applied in a range of fields such as feature selection, dimensionality reduction, text mining, and clustering \cite{b15,b16}. Paatero developed the NMF technique \cite{b17}, an unsupervised clustering algorithm that factorizes data matrices into two matrices by imposing non-negativity constraints on the components, one indicating the data partitions and the other representing the data set's cluster prototypes. The absence of negative values simplifies the interpretation of the generated matrices. For example, Brunet et al. \cite{b18} used NMF to identify distinctive molecular patterns. They demonstrated that NMF is more effective than other methods in obtaining biologically important information from cancer-related microarray data and is robust to initial settings.\par
\begin{figure}[t]
\includegraphics[width=\columnwidth]{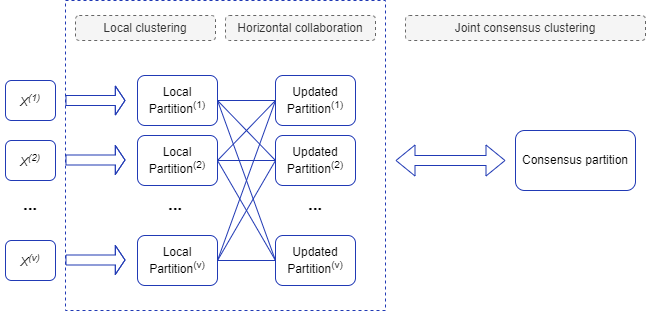}
\caption{Proposed pipeline for the joint multi-view collaborative clustering}
\label{fig:intro_jmvcc}
\end{figure}
The main difficulty in using NMF to multi-view clustering is determining how to restrict the search for factorizations to those that provide meaningful and comparable clustering solutions across several views at the same time. Furthermore, the NMF technique is non-convex algorithm and lacks solution unicity, which implies that if we run NMF on the same data set many times, we get different factorizations each time \cite{b28}.

Moreover, the existing studies on multi-view clustering either ignore the interaction between local partitions and the consensus partition or pay little attention to exchanging information between local partitions.\par
To address this gap, we set out to develop a novel approach, Joint Multi-View Collaborative Clustering (JMVCC), which allows information to be exchanged across multiple local partitions and seeks a consensus partition from these different views. Fig \ref{fig:intro_jmvcc} shows the pipeline of the proposed solution. First, we generate local partitions with the NMF method and improve the clustering quality of each view during the horizontal collaboration phase. Then, we combine these improved local partitions to form a consensus partition. Finally, using an iterative approach, the consensus partition leads the update of local partitions.\par
JMVCC provides three advantages over other multi-view clustering techniques:
\begin{itemize}
    \item The horizontal collaboration allows the local partitions to exchange information and enhance the quality of the clustering.
    \item The local and consensus partitions are mutually updated iteratively.
    \item During the fusion and horizontal collaboration phases, a weighting strategy is used to mitigate the risk of negative collaboration (i.e., collaboration between views with greater clustering quality and views with lower quality reduces the quality of the initial views), making the solution more robust to noisy views.
\end{itemize}
The rest of the paper is structured as follows: Section II goes through the preliminary setup as well as the formal definition of our solution JMVCC. Section III presents the solution optimization under various conditions. Finally, in Section IV, we evaluate the proposed algorithm's performance using various datasets. The paper ends with a conclusion and a discussion of several future works.

\section{Joint Multi-View Collaborative Clustering}
In this part, we introduce the NMF algorithm, followed by the formulation of our solution JMVCC and the subsequent functions.
\subsection{NMF algorithm}
Given a set of non-negative data matrices $X = \{X^{1}, X^{2},\cdots X^{V}\}$, such that $X^{v} = (x_1^v,x_2^v,\cdots, x_N^v) \in \mathbb{R}_+^{M\times N}$ is the data matrix corresponding to the $v$-th view, where $x_n^v \in \mathbb{R}_+^{M \times 1}$ represents the $n^{th}$ object of $X^{v}$, $v \in \mathbb{N}$.\par
The standard NMF algorithm factorizes the matrix $X^{v}$ into two low rank matrices $F^{v}$ and $G^{v}$, such that $F^{v}$ is the matrix of cluster centroids and $G^{v}$ is the matrix of data partitions defined respectively as $F^{v} = (f_1^v, f_2^v,\cdots,f_k^v) \in \mathbb{R}_+^{M\times K}$, and $G^{v} = (g_1^v, g_2^v,\cdots,g_N^v) \in \mathbb{R}_+^{K\times N}$, with $K$ a parameter representing the number of components. One of the NMF algorithm's frequent reconstruction methods can be expressed as a Frobenius norm optimization problem, which is stated as:
\begin{equation}\label{eq:std_nmf}
    \min_{F^{v},G^{v}} \mathcal{L}_v =  \Vert X^{v} - F^{v}G^{v} \Vert_F^2, \hspace{0.1cm} s.t.\hspace{0.1cm} F^{v} \geq 0, G^{v} \geq 0
\end{equation}
where $\Vert . \Vert_F$ represents the Frobenius norm and $F^{v} \geq 0, G^{v} \geq 0$ indicate the constraints that all matrices elements must be positive.

\subsection{Objective Function}
As previously stated, we want to use the horizontal collaboration method and the fusion of local partitions to enhance the clustering quality in a multi-view setting. To do so, we introduce the objective function of JMVCC, which is defined as follows:
\begin{align*}
\mathcal{J}(F^{v}, G^{v},G^*)&=\sum_{v=1}^{V}\left(\mathcal{L}_v(F^{v}, G^{v}) + \sum_{v'\neq v}^{V}\alpha_{v,v'} \hspace{0.1cm} \mathcal{H}(v,v')\right) \\
&+ \sum_{v}^{V}\beta_v\mathcal{U}_v(G^{v},G^*).
\end{align*}
The optimization problem can be written as:
\begin{equation}\label{eq:obj_fct}
\begin{split}
(F^{v},G^{v},G^*) &=  \argmin_{F^{v},G^{v},G^*}  \mathcal{J}(F^{v}, G^{v},G^*)
\end{split}
\end{equation}
where $\mathcal{L}_v(F^{v}, G^{v})$ is the standard NMF algorithm introduced in \eqref{eq:std_nmf}, $\mathcal{H}(v,v')$ is the multi-view collaboration term, $\mathcal{U}_v(G^{v},G^*)$ is the category utility function \cite{b19} between the consensus partition $G^*$ and the $v$-th local partition $G^{v}$, and $\alpha_{v,v'}$, $\beta_v$ are the weights of the horizontal collaboration and fusion respectively.\par 
The objective function combines consensus clustering with the generation and horizontal collaboration of local partitions to provide a one-step solution. In what follows, we define the horizontal collaboration term, the category utility function and the respective weights.\par
\subsection{Horizontal collaboration term}
A common collaborative strategy is to add information that has been retrieved from different views $v'\neq v$ to a view $v$ \cite{b19,b20}. In order to incorporate the information from several views, Grozavu et al. \cite{b20} introduced a multi-view collaboration technique that reduces the distance between a data point and its corresponding centroid.\par
Similarly, we want to include the information extracted from different views during the generation of the local partitions. In order to achieve this, we define the multi-view collaboration term as:
\begin{equation}\label{eq:fct_hcol}
    \mathcal{H}_{v,v'}(F^{v}, G^{v}) = \Vert F^{v}(G^{v} - G^{v'})\Vert_F^2.
\end{equation}
Our goal is to reduce the multi-view collaboration term using two data partition matrices of various views $G^{v'}$ and $G^{v}$. Notice that in \eqref{eq:fct_hcol} when $G^{v}$ and $G^{v'}$ agree, the collaborative term equals zero and we consider only the $v^{th}$ local NMF.

\subsection{Category Utility Function}
The category utility function is closely related to the contingency matrix, which measures the co-occurrence of two discrete random variables.\par
Given a consensus partition matrix $G^* \in \mathbb{R}^{K\times N}_+$ and a set of local partitions $G = \{G^{(1)}, G^{(2)},\cdots G^{V}\}$, we define the category utility function between $G^*$ and each $G^v$ with $ 1 \leq v \leq V$ as follows:
\begin{equation}\label{eq:utility_fct}
    \mathcal{U}(G^v,G^*) = \Vert G^v - G^* \Vert_F^2.
\end{equation}
Equation \eqref{eq:utility_fct} is simply the sum of the euclidean distances squared and is used to compute the discrepancy between the set of local partitions $G$ and the consensus partition matrix $G^*$.

\subsection{Weighting Strategy}
In order to minimize the reconstruction error in the solution proposed, we introduce weights during the horizontal collaboration and fusion phases: we investigate how optimizing the weights can lead to an optimal value of the global objective function \eqref{eq:obj_fct} and minimize the likelihood of negative collaboration by adjusting the weight factors during each iteration.\par
In \eqref{eq:obj_fct}, $\alpha_{v,v'}$ is the weight of the horizontal collaboration between the $v$-th view and the other views $v'\neq v$ and $\beta_v$ is the weight used during the fusion phase. Additionally, the weights $\alpha$ and $\beta$ are learnable parameters such that $\sum_{v'\neq v}\alpha_{v,v'}^\gamma = 1$ and $\sum_v\beta_v^\gamma = 1$, with $\gamma > 1$ a parameter that controls the weights distribution. Hence, important views (i.e., views with better clustering quality) will receive higher weights during the update.

\section{Optimization Algorithm}
Given that the Frobenius norm is continuous and differentiable, it follows that the cost function given in \eqref{eq:obj_fct} is also differentiable, and its derivative exists at every point in its domain. As a result, there is always a minimum. However, the NMF algorithm is non-convex in both $F^{v}$ and $G^{v}$, therefore it is not possible to find the global minimum.\par
To overcome the problem of non-uniqueness of the NMF algorithm, we proposed the use of the aforementioned weighting strategy.\par 
Moreover, to minimize the objective function defined in \eqref{eq:obj_fct}, we propose a convergent iterative update approach based on the gradient descent optimization technique. Specifically, we consider Lee and Seung’ technique \cite{b21} and employ an adaptive learning rate to ensure that the non-negativity constraint is satisfied, as well as the multiplicative update rule for the centroid, partition and consensus matrices $F$, $G$ and $G^*$ during optimization. This implies that for $\Theta \in \{F^{v}, G^{v}, G^*\}$ $st.\hspace{0.2cm} \Theta \geq 0$, the adaptive learning rate $\eta_\Theta$ is written as:
\begin{equation}\label{eq:adapt_lr}
    \eta_\Theta = \frac{\Theta}{[\nabla_\Theta \mathcal{J}]_+}
\end{equation}
and the multiplicative update rule is defined as:
\begin{equation}\label{eq:multi_update}
\Theta = \Theta \circ \frac{[\nabla_{\Theta}\mathcal{J}]_-}{[\nabla_{\Theta}\mathcal{J}]_+}.
\end{equation}
Here, $\mathcal{J}$ is the functional defined in \eqref{eq:obj_fct}, the fraction line and $\circ$ represent the element-wise division and the Hadamard product respectively. $[.]_+$ and $[.]_-$ are the positive and negative terms of the gradient respectively.\par

\subsection{Iterative Update Approach}
The proposed iterative update technique is described in detail in the paragraphs that follow.
\subsubsection{\textbf{Fixing $G^*$ and $F^{v}$, minimize \eqref{eq:obj_fct} over $G^{v}$}}
When $G^*$ and $F^{v}$ are fixed, the optimization leads to the following update rule:
\begin{equation}\label{eq:partition_update}
    G^v = G^v \circ \frac{{F^v}^TX^v + \sum_{v'\neq v}\alpha_{v,v'} \left ({F^v}^TF^vG^{v'} \right ) + \beta_vG^*}{{F^v}^TF^vG^v + \sum_{v'\neq v}\alpha_{v,v'} \left ({F^v}^TF^vG^v \right ) + \beta_vG^v}.
\end{equation}

\subsubsection{\textbf{Fixing $G^*$ and $G^{v}$, minimize \eqref{eq:obj_fct} over $F^{v}$}}
When $G^*$ and $G^{v}$ are fixed, we only consider the standard NMF and horizontal collaboration functions. The following updating rule results from the minimization:
\begin{equation}\label{eq:basis_update}
    F^v = F^v \circ \frac{X^v{G^v}^T + \sum_{v'\neq v}\alpha_{v,v'} \left (F^vG^v{G^{v'}}^T+G^{v'}{G^v}^T\right )}
    {F^vG^v{G^v}^T + \sum_{v'\neq v}\alpha_{v,v'} \left (F^vG^v{G^v}^T + G^{v'}{G^v}^T\right )}.
\end{equation}

\subsubsection{\textbf{Fixing $G^{v}$ and $F^{v}$, minimize \eqref{eq:obj_fct} over $G^*$}} Here, we only consider the consensus term in our objective function. The update rule for the consensus partition matrix $G^*$ is defined as:
\begin{equation}\label{eq:consensus_update}
    G^* = G^* \circ \frac{\sum_{v}^{V}\beta_v G^{v}}{\sum_{v}^{V}\beta_v G^*}.
\end{equation}
Note that $G^*$ is also positive by definition.

\subsection{Update of the horizontal collaboration weight}
Our goal is to minimize the objective function defined in \eqref{eq:obj_fct} with respect to $\alpha_{v,v'}$. We follow the strategy proposed by Cai et al. \cite{b22} and only consider the horizontal collaboration term
\begin{equation}\label{eq:weight_hcol}
    \min_{\alpha_{v,v'}} \sum_{v'\neq v}\alpha_{v,v'} \mathcal{H}_{v,v'}, \hspace{0.2cm} s.t., \sum_{v'\neq v}\alpha^\gamma_{v,v'} = 1,\hspace{0.1cm} \alpha_{v,v'} \geq 0,
\end{equation}
where $\mathcal{H}_{v,v'}$ is the collaboration function defined in \eqref{eq:fct_hcol}. Taking into account the aforementioned constraints, we define the Lagrangian function of \eqref{eq:weight_hcol} as:
\begin{equation}\label{eq:lagr_fct}
    \sum_{v'\neq v}\alpha_{v,v'}  \mathcal{H}_{v,v'} - \lambda_{v,v'} \hspace{0.1cm} \left (\sum_{v'\neq v}\alpha^\gamma_{v,v'} - 1  \right ) - \theta_{v,v'} \hspace{0.1cm} \alpha_{v,v'}
\end{equation}
where $\lambda_{v,v'}$ and $\theta_{v,v'}$ are Lagrange multipliers, with respect to ${v',v}$.
By setting the derivative of \eqref{eq:lagr_fct} with respect to $\alpha_{v,v'}$ to zero, then substituting the resulting $\alpha_{v,v'}$ into the constraint $ \sum_{v'\neq v}\alpha_{v,v'}^\gamma = 1$, we get:
\begin{equation}\label{eq:alpha_fct}
    \alpha_{v,v'} = \frac{(\mathcal{H}_{v,v'})^{\frac{1}{\gamma - 1}}}{\sum_{v'\neq v} (\mathcal{H}_{v,v'})^{\frac{1}{\gamma - 1}}}.
\end{equation}

\subsection{Update of the consensus weight}
We present another method for optimizing the consensus weights based on the Karush-Kuhn-Tucker (KKT) conditions \cite{b23}.\par
Since $\beta_v \geq 0$, our goal is to find the positive weights that will determine the contribution strength of each local partition $G^{v}$ to the consensus partition $G^*$. Additionally, and following the update of the matrices $F^v$, $G^v$ and $G^*$, we only consider the consensus function $\mathcal{U}_v$ defined in \eqref{eq:utility_fct} in order to optimize the consensus weights $\beta_v$. In what follows, we define $\phi$ as the consensus functional:
\begin{equation}
    \begin{cases}
    \beta_v = \argmin_\beta \phi = \argmin_\beta \sum_v^V \beta _v\hspace{0.1cm} \mathcal{U}_v(G^v,G^*) \\
    \forall v \in V \hspace{0.2cm} \sum_v^V \beta^\gamma_v = 1 \hspace{0.2cm} and \hspace{0.2cm} \beta_v \geq 0.
    \end{cases}
\end{equation}
Using Lagrange multipliers $\lambda_v$ and $\theta_v$ for the above system, we get the following KKT conditions, for $v \in V$:
\begin{equation}
    \begin{cases}
    \beta_v  \geq 0 \\
    \sum_v^V \beta^\gamma_v = 1 \\
    \lambda \cdot \beta_v  = 0 \\
    \nabla \phi_v - \lambda_v \nabla \beta_v  + \theta_v \nabla (\beta^\gamma_v - 1) = 0.
    \end{cases}
\end{equation}
Below are the optimization's results under the Karush-Kuhn-Tucker conditions:
\begin{equation}\label{eq:beta_fct}
    \beta_v = \frac{\left(\mathcal{U}_v(G^v,G^*)\right)^{\frac{1}{\gamma - 1}}}{\sum_v \left(\mathcal{U}_v(G^v,G^*)\right)^{\frac{1}{\gamma - 1}}}.
\end{equation}
Notice that all the horizontal collaboration and consensus terms receive equal weight factors when $\gamma \rightarrow \infty$. Moreover, we avoid the trivial solution of $\gamma \rightarrow 1$. Finally, the formulation in \eqref{eq:alpha_fct} and \eqref{eq:beta_fct} enables us to regulate every weight factor with just one parameter $\gamma$.\par
The proposed solution is summarised in Alg. \ref{alg:algoG}.

\begin{algorithm}[H]
\SetAlgoLined
\label{alg:algoG}
\vspace{0.05cm}
\caption{JMVCC Algorithm}

\vspace{0.05cm}
\textbf{Input:} Multi-view data set $X = \{X^{1}, X^{2}, \cdots X^{v}\}$, with $X^{v} \in \mathbb{R}_+^{M \times N}$ and $v$ number of views \newline
The number of components $K$ \newline
The parameter $\gamma$\\
\textbf{Initialization:} Randomly initialize the consensus matrix $G^*$ and the matrices $F^{v}$ and $G^{v}$\\
\textbf{For all realizations} \\
      \ForAll{views $v$ } {
      	Compute the optimized $\alpha_{v,v'}$ with \eqref{eq:alpha_fct}. \\
            Compute the optimized $\beta_v$ with \eqref{eq:beta_fct}.\\
		Update the partition matrices of all views  \eqref{eq:partition_update}. \\ 	 
		Update the centroid matrices of all views  \eqref{eq:basis_update}. \\
        Update the consensus matrix with \eqref{eq:consensus_update}
 	}
	 	  
\end{algorithm}

\section{Experiments}
In this part, we evaluate the performance of our suggested joint multi-view collaborative clustering solution using three multi-view data sets, namely NUS-WIDE, Caltech101 and Handwritten. To better explain the premise of the proposed approach, more information about the data sets is provided. Moreover, Since the labels are made available for these data sets, we use two metrics to assess the results of our solution: the purity score and the Normalized Mutual Information (NMI).

\subsection{Data Sets Description}
\begin{itemize}
    \item \textbf{NUS-WIDE\footnote{https://lms.comp.nus.edu.sg/wp-content/uploads/2019/research/nuswide/NUS-WIDE.html} -} it includes 5,018 distinct tags and 269,648 images from Flickr, which correspond to 81 classes. Additionally, six views are provided: the color histogram, color correlogram, edge direction histogram, wavelet texture, block-wise color moments, and a bag of visual words on SIFT descriptions. In order to conduct our experiment, we extracted two subsets (NUS-2B and NUS-CDF); (see Tab.\ref{tab:datasets}).
    \item \textbf{Caltech101\footnote{https://github.com/yeqinglee/mvdata} - } it contains 101 categories represented by 8677 photos. We selected the seven most popular categories: Faces, Motorbikes, Dolla-Bill, Garfield, Snoopy, Stop-Sign, and Windsor Chair, which is equivalent to 1474 images. Following \cite{b24}, we extract six views, namely Gabor features, Wavelet Moments (WM), CENTRIST, HOG, GIST, LBP. Tab. \ref{tab:datasets} gives more details about these views. 
    \item \textbf{Handwritten\footnote{https://mvlearn.github.io/references/datasets.html} - } it consists of 2000 data samples for ten digits (0 to 9) with 200 data samples per class/digit. We test our solution on two views. Specifically, we use the pixel averages in $2\times 3$ windows (PIX) and the Fourier coefficients of the digit shapes (FOU).
\end{itemize}
\begin{table}[t]
\caption{Experimental Data sets}
\label{tab:datasets}
\resizebox{\columnwidth}{!}{%
\def\arraystretch{1.5}%
\begin{tabular}{|c|c|c|c|}
\hline
\textbf{Data sets}    & \textbf{\# Instances} & \textbf{\# Clusters} & \textbf{\# Features}                  \\ \hline
NUS-CDF      & 3930        & 3          & 64, 144, 75, 128, 225, 500  \\ \hline
NUS-2B       & 4701        & 2          & 64, 144, 75, 128, 225, 500  \\ \hline
Caltech101-7 & 1474        & 7          & 48, 40, 254, 1984, 512, 928 \\ \hline
Handwritten  & 2000        & 10         & 240, 74                     \\ \hline
\end{tabular}%
}
\end{table}

\begin{table*}[t]
\caption{Clustering performance on four data sets}
\label{tab:scores}
\centering
\def\arraystretch{1.5}%
\begin{tabular}{|l|llll|llll|}
\hline
\multicolumn{1}{|c|}{\multirow{2}{*}{\textbf{Algorithms}}} & \multicolumn{4}{c|}{\textbf{Purity score (\%)}}                                                                             & \multicolumn{4}{c|}{\textbf{NMI score (\%)}}                                                                                \\ \cline{2-9} 
\multicolumn{1}{|c|}{}                            & \multicolumn{1}{l|}{Caltech101-7} & \multicolumn{1}{l|}{Handwritten}     & \multicolumn{1}{l|}{NUS-CDF}    & NUS-2B     & \multicolumn{1}{l|}{Caltech101-7} & \multicolumn{1}{l|}{Handwritten}     & \multicolumn{1}{l|}{NUS-CDF}    & NUS-2B     \\ \hline
ColNMF                                            & \multicolumn{1}{l|}{60.7 $\pm$ 2.0}   & \multicolumn{1}{l|}{51.9 $\pm$ 1.3} & \multicolumn{1}{l|}{34.1 $\pm$ .0}  & 39.2 $\pm$ .05 & \multicolumn{1}{l|}{49.1 $\pm$ 2.6}   & \multicolumn{1}{l|}{50.3 $\pm$ 1.6} & \multicolumn{1}{l|}{21.9 $\pm$ 0.8} & 38.1 $\pm$ .2  \\ \hline
C-NMF                                             & \multicolumn{1}{l|}{70.8 $\pm$ .05}   & \multicolumn{1}{l|}{64.8 $\pm$ .01} & \multicolumn{1}{l|}{51.9 $\pm$ .08} & 67.6 $\pm$ .02 & \multicolumn{1}{l|}{70.7 $\pm$ .0}    & \multicolumn{1}{l|}{50.7 $\pm$ .02} & \multicolumn{1}{l|}{43.2 $\pm$ .1}  & 64.3 $\pm$ .02 \\ \hline
Multi-NMF                                         & \multicolumn{1}{l|}{58.9 $\pm$ .01}   & \multicolumn{1}{l|}{74.2 $\pm$ 2.1} & \multicolumn{1}{l|}{46.3 $\pm$ .0}  & 52.7 $\pm$ .0  & \multicolumn{1}{l|}{56.1 $\pm$ .1}    & \multicolumn{1}{l|}{71.6 $\pm$ .01} & \multicolumn{1}{l|}{44.1 $\pm$ .0}  & 45.0 $\pm$ .0  \\ \hline
CMVC                                              & \multicolumn{1}{l|}{67.6 $\pm$ .0}    & \multicolumn{1}{l|}{81.8 $\pm$ .0}  & \multicolumn{1}{l|}{61.5 $\pm$ .0}  & 63.6 $\pm$ .0  & \multicolumn{1}{l|}{64.3 $\pm$ .0}    & \multicolumn{1}{l|}{77.8 $\pm$ .0}  & \multicolumn{1}{l|}{58.2 $\pm$ .0}  & 62.3 $\pm$ .0  \\ \hline
JMVCC                                      & \multicolumn{1}{l|}{\textbf{78.0 $\pm$ .02}}             & \multicolumn{1}{l|}{\textbf{83.4 $\pm$ .04}}           & \multicolumn{1}{l|}{\textbf{89.8 $\pm$ .01}}           &     \textbf{80.5 $\pm$ .2 }      & \multicolumn{1}{l|}{\textbf{76.5 $\pm$ .03}}             & \multicolumn{1}{l|}{\textbf{77.9 $\pm$ .04}}           & \multicolumn{1}{l|}{\textbf{83.0 $\pm$ .0}}           &    \textbf{76.2 $\pm$ .03}        \\ \hline
\end{tabular}
\end{table*}

\subsection{Evaluation Metrics}
Since the ground truth of the data sets is readily available, we evaluate our approach with the commonly-used external measures Purity score and NMI for cluster validity.

\subsubsection{\textbf{Purity score}} it assesses the degree to which a cluster belongs to a class. Let $P = \{p_1, p_2, \cdots p_n\}, n \in \mathbb{N}$ and $K = \{k_1, k_2, \cdots k_m\}, m \in \mathbb{N}$ be the known data labels and centroids respectively. The purity score is defined as:
\begin{equation}
    purity = \frac{1}{N}\sum_{m=1}^{|K|} 
\max_{i=1}^{|P|} |k_m\cap p_n|.
\end{equation}
The clustering result's purity is equal to the expected purity of all clusters. A purity of 1 implies good clustering, while a purity of 0 implies poor clustering.

\subsubsection{\textbf{Normalized Mutual Information}} NMI is a version of Mutual Information, a common measure in information theory. It is frequently regarded because of its ability to compare two partitions with varied numbers of clusters. It is defined as:
\begin{equation}
    NMI(P,K) = \frac{2 \times I(P;K)}{H(P) + H(K)},
\end{equation}
where $I(P;K)$ is the mutual information between the ground truth $P$ and the clustered set $K$. $H(P)$ and $H(K)$ represent the entropy.

\subsection{Baseline Algorithms}
We compare the proposed Joint Multi-View Collaborative Clustering approach against a number of baseline algorithms for multi-view and ensemble clustering to demonstrate its utility.
\begin{itemize}
    \item \textbf{ColNMF - } it uses a shared coefficient matrix with various basis matrices across views to handle multi-view data \cite{b25}, as seen below:
    \begin{equation}
        \sum_v^V \alpha_v \Vert X^{v} - F^{v}{G^*}^T \Vert_F^2.
    \end{equation}
    \item \textbf{C-NMF - } introduced by Benlamine et al. \cite{b26}, it is divided into two steps. The approach first learns a NMF for a fixed number of components K for each view, then computes a consensus solution for the various NMF models in the second step.
    \item \textbf{MultiNMF - } it seeks a factorization that produces compatible clustering solutions across many views \cite{b2}. It is done by developing a collaborative matrix factorization process with the constraint of pushing each clustering solution towards a consensus rather than directly fixing it.
    \item \textbf{CMVC - } it generates several basic partitions from each view, followed by ensemble clustering to obtain the consensus partition; the later updates the basic partitions jointly and iteratively \cite{b27}.
\end{itemize}

\subsection{Results}
The clustering performance of various methods across all four data sets is presented in Tab. \ref{tab:scores}. 10 test runs with various random initializations were performed, and the results are provided along with the average performance and standard deviation.\par
Since the consensus matrix $G^*$ is a continuous matrix, we discretize it by computing the predominant basis component. The row index for which the item is the maximum within the column is used to determine the dominating basis component.\par
In terms of purity/normalized mutual information, JMVCC exceeds the second best algorithm with a gap of $7.2$\%/$5.8$\% on Caltech101, $1.6$\%/$1.1$\% on Handwritten, $28.3$\%/$24.8$\% on NUS-CDF and $12.9$\%/$11.9$\% on NUS-2B.\par
On the NUS-CDF and NUS-2B data sets, JMVCC significantly outperforms the baseline approaches with an average of $20.6$\%/$18.3$\%. This can be explained by the presence of noisy views in these data sets, which are assigned less weight than views with higher clustering quality, thus minimizing the impact of these unfavorable views. The proposed JMVCC can effectively find high-quality consensus clustering because it reduces the impact of noisy views by implementing a weighting strategy during both the horizontal collaboration and fusion steps. As a result, it confirms our claim that the proposed approach JMVCC is quite useful in clustering multi-view data sets with noisy views.\par
Both JMVCC and CMVC show promising clustering results on the Caltech101-7 and Handwritten digit data sets. JMVCC performs slightly better than CMVC and gets a $14.5$\%/$19.8$\% performance advantage over the other algorithms. JMVCC improves its clustering solution by exchanging extracted knowledge from diverse views $v \in V$ during the horizontal collaboration phase.

\begin{figure*}[t]
\centering
\includegraphics{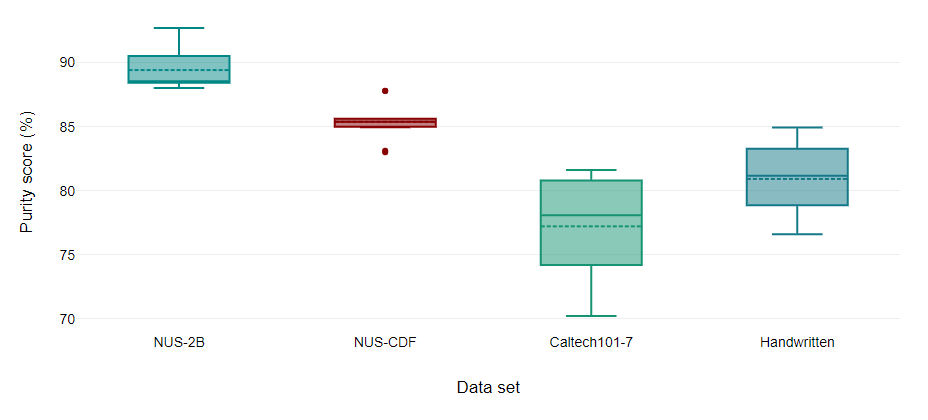}
\caption{Comparative results on different data sets.}
\label{fig:box_plot}
\end{figure*}

\begin{figure}[t]
\centering
\includegraphics[width=\columnwidth]{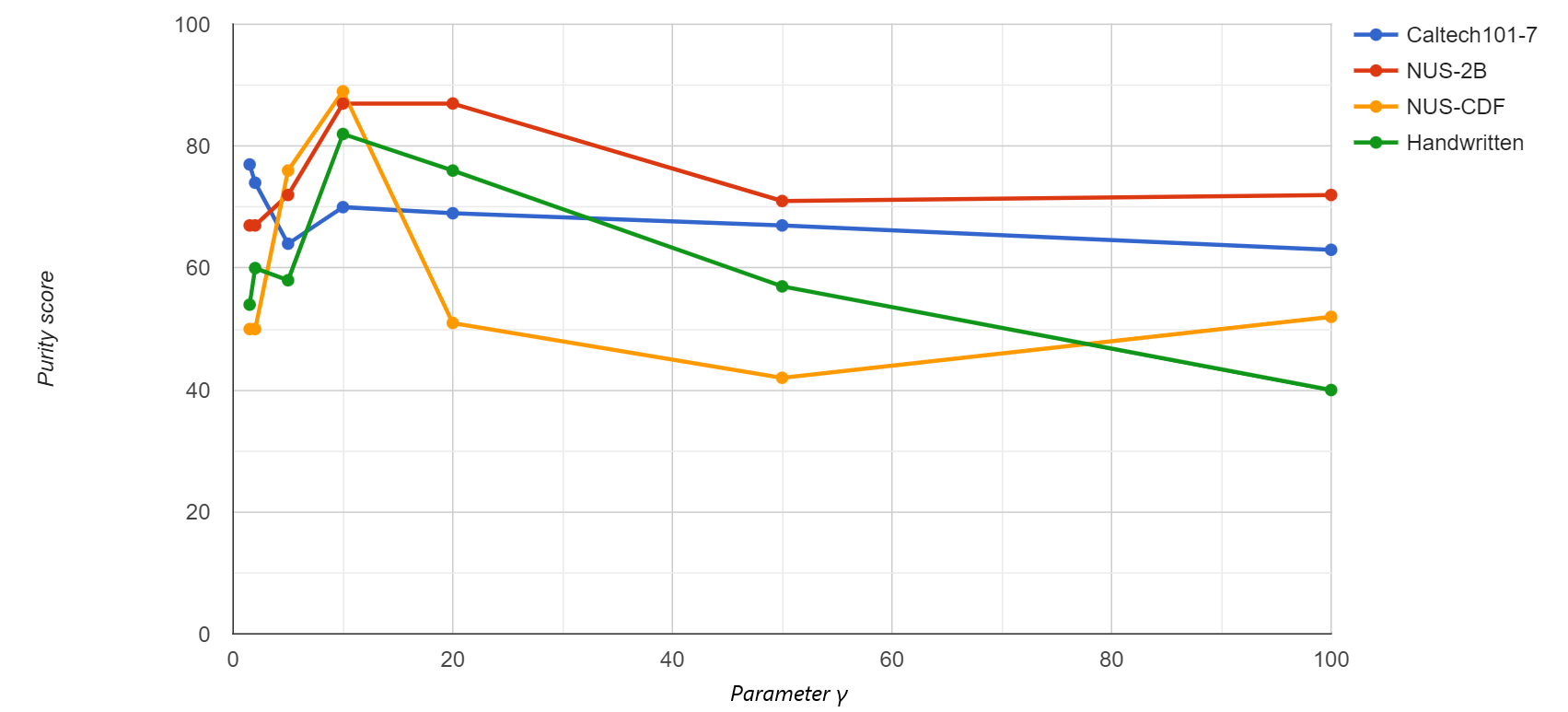}
\caption{Impact of the integer parameter $\gamma \geq 2$ on the purity score.}
\label{fig:gamma_impact}
\end{figure}

\subsection{Parameter study}
If each individual view gives higher weights to views that have the most similar solutions to the local one in the setting of horizontal collaboration and fusion, the global results are expected to be better.\par 
Further investigation reveals, in Fig. \ref{fig:gamma_impact}, that the degree to which one view $v$ should collaborate with other views $v' \neq v$ with dissimilar partitions is determined by the parameter $\gamma$. If multiple views have similar clustering partitions, they will be given equal weight. The views with the most similar partitions would still be preferred to optimize the cost function of the proposed solution even while using a larger $\gamma$. However, views with less identical partitions would still be taken into account. In fact, the weight of the partitions from different views would increase as $\gamma$ increases, eventually equaling the weight of all other partitions. When the value of $\gamma$ is high enough in this later scenario, it would be comparable to giving each view the same weight.

\section{Conclusion}
In this paper, we presented a novel technique for multi-view consensus clustering between different local Non-negative Matrix Factorizations. The suggested approach uses the horizontal collaboration technique to update the local clusterings associated with each local view, fusing these updated clusterings into a consensus clustering which then iteratively and alternately updates local clusterings. Additionally, we introduced a weighting strategy to mitigate the risk of negative information exchange during the horizontal collaboration and consensus steps.\par 
The results of the experiments demonstrate that the suggested method, which has been verified against a number of data sets, produces better results with respect to the standard multi-view consensus clustering methods in terms of clustering quality. Furthermore, we show that the proposed technique is only barely influenced by random initialization.\par
As part of our future work, we plan to develop a strategy in the case of $\gamma \rightarrow 1$ where each view $v$ would only collaborate with the views $v' \neq v$ that have the most similar solution. We also plan to implement the solution proposed in a deep learning context using the Multi-Layer NMF. Finally, we will integrate white noise and conduct extensive comparisons with other methods to properly assess the robustness of the proposed solution.


\begin{thebibliography}{00}
\bibitem{b1} Yan, A., Wang, W., Ren, Y. and Geng, H., 2021. A Clustering Algorithm for Multi-Modal Heterogeneous Big Data With Abnormal Data. Frontiers in Neurorobotics, 15, p.64.
\bibitem{b2} Liu, J., Wang, C., Gao, J. and Han, J., 2013, May. Multi-view clustering via joint nonnegative matrix factorization. In Proceedings of the 2013 SIAM international conference on data mining (pp. 252-260). Society for Industrial and Applied Mathematics.
\bibitem{b3} Sun, S., 2013. A survey of multi-view machine learning. Neural computing and applications, 23(7), pp.2031-2038.
\bibitem{b4} Eaton, E., Desjardins, M. and Jacob, S., 2010, October. Multi-view clustering with constraint propagation for learning with an incomplete mapping between views. In Proceedings of the 19th ACM international conference on Information and knowledge management (pp. 389-398).
\bibitem{b5} Kim, Y.M., Amini, M.R., Goutte, C. and Gallinari, P., 2010, July. Multi-view clustering of multilingual documents. In Proceedings of the 33rd international ACM SIGIR conference on Research and development in information retrieval (pp. 821-822).
\bibitem{b6} Bickel, S. and Scheffer, T., 2004, November. Multi-view clustering. In ICDM (Vol. 4, No. 2004, pp. 19-26).
\bibitem{b7} Wang, S., Ye, Y. and Lau, R.Y., 2015, August. A Generative Model with Ensemble Manifold Regularization for Multi-view Clustering. In International Conf. on Intelligent Computing (pp. 109-114). Springer.
\bibitem{b8} Wei, X., Cao, B. and Philip, S.Y., 2017, May. Multi-view unsupervised feature selection by cross-diffused matrix alignment. In 2017 International Joint Conference on Neural Networks (IJCNN) (pp. 494-501). IEEE.
\bibitem{b9} Kumar, A., Rai, P. and Daume, H., 2011. Co-regularized multi-view spectral clustering. Advances in neural information processing systems, 24.
\bibitem{b10} Yang, Y. and Wang, H., 2018. Multi-view clustering: A survey. Big Data Mining and Analytics, 1(2), pp.83-107.
\bibitem{b11} De Sa, V.R., Gallagher, P.W., Lewis, J.M. and Malave, V.L., 2010. Multi-view kernel construction. Machine learning, 79(1), pp.47-71.
\bibitem{b12} Wang, X., Qian, B., Ye, J. and Davidson, I., 2013, May. Multi-objective multi-view spectral clustering via pareto optimization. In Proceedings of the 2013 SIAM international conference on data mining (pp. 234-242). Society for Industrial and Applied Mathematics.
\bibitem{b13} Khalafaoui, Y., Grozavu, N., Matei, B. and Goix, L.-W., 2022. Multi-modal Multi-view Clustering Based on Non-negative Matrix Factorization. In press: 2022 IEEE Symposium Series on Computational Intelligence (IEEE SSCI). pp.1386–1391.
\bibitem{b14} Jiang, Y., Liu, J., Li, Z., Li, P. and Lu, H., 2013. Co-regularized PLSA for multi-view clustering. In Asian Conference on Computer Vision (pp. 202-213). Springer, Berlin, Heidelberg.
\bibitem{b15} Cichocki, A., Zdunek, R., Phan, A.H. and Amari, S.I., 2009. Nonnegative matrix and Tensor Factorizations: Applications to Exploratory Multi-way Data Analysis and Blind Source Separation. John Wiley \& Sons.
\bibitem{b16} Kim, J. and Park, H., 2008. Sparse Nonnegative Matrix Factorization for Clustering. Georgia Institute of Technology.
\bibitem{b17} Paatero, P. and Tapper, U., 1994. Positive Matrix Factorization: A Non‐negative Factor Model with Optimal Utilization of Error Estimates of Data Values. Environmetrics, 5(2), pp.111-126.
\bibitem{b18} Brunet, J.P., Tamayo, P., Golub, T.R. and Mesirov, J.P., 2004. Metagenes and molecular pattern discovery using matrix factorization. Proceedings of the national academy of sciences, 101(12), pp.4164-4169.
\bibitem{b19} Mirkin, B., 2001. Reinterpreting the category utility function. Machine Learning, 45(2), pp.219-228.
\bibitem{b20} Grozavu, N., Matei, B., Bennani, Y. and Benlamine, K., 2022. Multi-view Clustering Based on Non-negative Matrix Factorization. In Recent Advancements in Multi-View Data Analytics (pp. 177-200). Springer.
\bibitem{b21} Seung, D. and Lee, L., 2001. Algorithms for Non-negative Matrix Factorization. Advances in neural information processing systems, 13, pp.556-562.
\bibitem{b22} Cai, X., Nie, F. and Huang, H., 2013, June. Multi-view k-means clustering on big data. In Twenty-Third International Joint conference on artificial intelligence.
\bibitem{b23} H. W. Kuhn and A. W. Tucker. Nonlinear Programming. In Berkeley University of California Press, editor,Proceedings of 2nd Berkeley Symposium, pages 481–492,1951.
\bibitem{b24} Li, Y., Nie, F., Huang, H. and Huang, J., 2015, February. Large-scale multi-view spectral clustering via bipartite graph. In Twenty-ninth AAAI conference on artificial intelligence.
\bibitem{b25} Singh, A.P. and Gordon, G.J., 2008, August. Relational learning via collective matrix factorization. In Proceedings of the 14th ACM SIGKDD international conference on Knowledge discovery and data mining (pp. 650-658).
\bibitem{b26} Benlamine, K., Bennani, Y., Matei, B., Grozavu, N. and Falih, I., 2022. Collaborative Learning to Improve the Non-uniqueness of NMF. International Journal of Computational Intelligence and Applications, 21(01), p.2250001.
\bibitem{b27} Liu, H. and Fu, Y., 2018. Consensus guided multi-view clustering. ACM Transactions on Knowledge Discovery from Data (TKDD), 12(4), pp.1-21.
\bibitem{b28} Wang, Y.X. and Zhang, Y.J., 2012. Nonnegative matrix factorization: A comprehensive review. IEEE Transactions on knowledge and data engineering, 25(6), pp.1336-1353.
\end{thebibliography}
\end{document}